\pdfoutput=1
\documentclass{article}
\PassOptionsToPackage{numbers, compress}{natbib}
\usepackage[preprint]{neurips_2021}
\usepackage[utf8]{inputenc} % allow utf-8 input
\usepackage[T1]{fontenc}    % use 8-bit T1 fonts
\usepackage{url}            % simple URL typesetting
\usepackage{booktabs}       % professional-quality tables
\usepackage{amsfonts}       % blackboard math symbols
\usepackage{nicefrac}       % compact symbols for 1/2, etc.
\usepackage{microtype}      % microtypography
\usepackage{xcolor}         % colors
\usepackage{amsmath, amsthm}
\usepackage{graphicx}
\usepackage{mathtools}
\usepackage{amsmath,amsfonts,amssymb}
\usepackage{wrapfig}
\usepackage{subfig}
\usepackage{multirow}
\usepackage{multicol}
\title{A generative model for molecule generation based on chemical reaction trees}

\author{%
  Dai Hai Nguyen\\
  The University of Tokyo\\
  \texttt{hai@k.u-tokyo.ac.jp} \\
   \And
   Koji Tsuda \\
  The University of Tokyo\\
  \texttt{tsuda@k.u-tokyo.ac.jp} \\
}
\begin{document}

\maketitle

\begin{abstract}
  Deep generative models have been shown powerful in generating novel molecules with desired chemical properties via their representations such as strings, trees or graphs.
  However, these models are limited in recommending synthetic routes for the generated molecules in practice.
  We propose a generative model to generate molecules via \textit{multi-step chemical reaction trees}. Specifically, our model first propose a chemical reaction tree with predicted reaction templates and commercially available molecules (starting molecules), and then perform forward synthetic steps to obtain product molecules.
  Experiments show that our model can generate chemical reactions whose product molecules are with desired chemical properties. Also, the complete synthetic routes for these product molecules are provided.
\end{abstract}

\section{Introduction}
Finding small molecules with desired chemical properties is a key challenge in drug discovery. A significant amount of recent research work has explored the use of deep generative models such as variational autoencoders (VAEs) \cite{kingma2013auto} or generative adversarial networks (GANs) \cite{goodfellow2014generative} for generating small molecules. These models used either SMILES based \cite{gomez2018automatic,kusner2017grammar} or graph based \cite{jin2018junction,bowman2015generating} representations, and can produce chemically valid molecules with desired properties effectively.

However, these methods do not account for synthetic feasibility of generated molecules, i.e. describing how to synthesize them using a given set of commercially available reactants (also called starting molecules). To handle this issue, MOLECULE CHEF \cite{bradshaw2019model} attempts to generate a bag of reactants using a generative model, then predict a product molecule through a reaction predictor. Even though MOLECULE CHEF allows to simultaneously generate molecules with desired chemical properties and suggest how they can be made by the predicted reactants, it is restricted to molecules which are products of single-step chemical reactions, which leads to the limited diversity of generated molecules. 

\begin{wrapfigure}[32]{r}{0.44\textwidth}
  \begin{center}
    \includegraphics[width=0.44\textwidth]{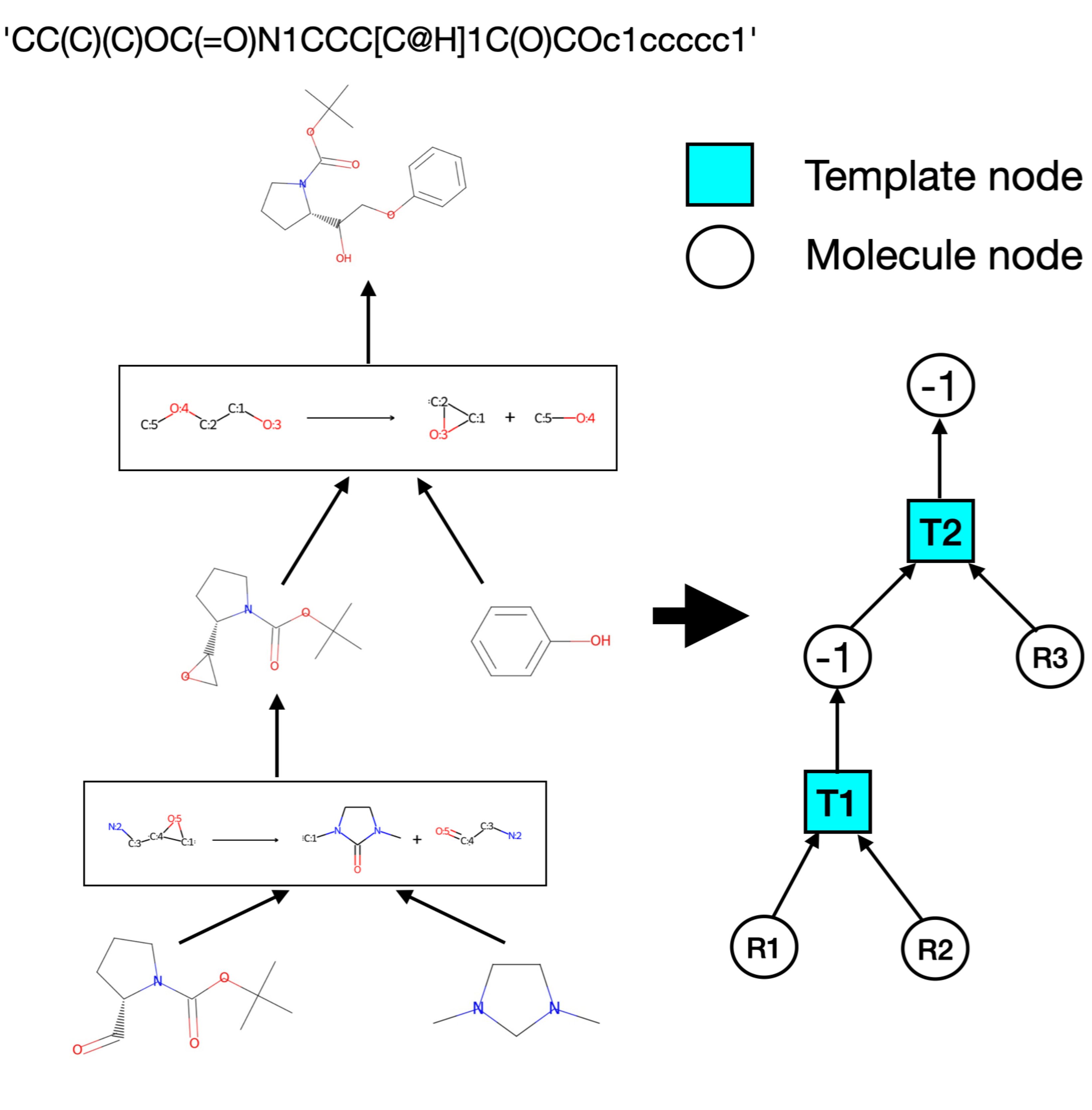}
  \end{center}
  \caption{Example of a synthetic route generated by our model and represented by a reaction tree, in which circle nodes represent molecules and rectangles represent reaction templates. The template and starting molecules nodes are labeled by their indices in the vocabularies of reaction templates and starting molecules, while molecule nodes of the root and intermediaries are labeled -1. The final product molecule at the root is obtained by performing chemical transformation steps from the starting molecule nodes towards the root.}
  \label{Fig:chemicalreactiontree}
\end{wrapfigure}

In this paper, we explore an alternative framework that learn to generate \emph{multi-step chemical reactions}, which are represented by tree-like structures, called (chemical) reaction trees \cite{shibukawa2020compret}, where molecule and reaction template nodes alternatively appear, as shown in Figure \ref{Fig:chemicalreactiontree}. To this end, we propose a generative model that generates synthetic routes with reaction templates and available starting molecules in three phases by exploiting valid chemical substructures given in a junction tree \cite{jin2018junction}. First, the model generate a junction tree, which has a set of substructures and their relative arrangement. Second, a reaction tree is generated with the guide from the generated junction tree. Third, the reaction tree is required to be valid in the sense that we can perform a number of chemical transformation steps from the starting molecule nodes towards the root, where the final product molecule is obtained.

%To this end, we propose a generative model over chemical reaction trees with the following architecture: 1) \textit{encoder}: learning to map from a chemical reaction tree to continuous latent space in the bottom-up fashion; 2) \textit{decoder}: learning to reconstruct the chemical reaction tree from the latent vector in latent space in the top-down fashion. In the generation process, we sample a latent vector from a given prior distribution and then construct a chemical reaction tree. The generated reaction tree is required to be valid in the sense that we can perform a number of forward synthesis steps with predicted reactants and templates towards the root. The final product molecule is obtained at the root node (final step). 
%, from which we can generate the product molecules. Specifically, the model has two parts:
%In this paper, we tackle two problems of molecule generation: (1) generation of valid molecules with desired chemical properties and (2) recommendation of synthetic routes (or tree like structure of multi-step chemical reactions) for generated molecules from a set of starting reactants and templates from literature. 
%To this end, we propose a generative model over chemical reaction trees with the following architecture:
%Thus, our model not only generates molecules which are resulted from multi-step chemical reactions, but also give associated chemical reaction trees for synthesizing them. This novel point has not addressed in the previous methods to our best knowledge. 

To summarize, our model has several advantages over previous ones: 1) \textit{Synthetizability}: each molecule generated by our model is associated with a reaction tree where reactants and a reaction template for each reaction step are provided; 2) \textit{Desired chemical properties}: the model can produce novel molecules with desired properties of interest by performing Bayesian optimization on the latent representations of their associated reaction trees. While MOLECULE CHEF is template-free and heavily reliant on the reaction predictor to obtain the product molecule, our model is template-based and does not need any additional module. Especially, the model can also deal with multi-step chemical reactions, thus providing the complete synthetic routes for each generated molecule.

\section{Related work}
In this work, we approach the task of molecular generation by taking advantage of generative models. To highlight the improvements of our proposed method, in this section, we focus on the methodologies which adopted generative models to generate small molecules. The common scheme has two parts: 1) \textit{encoding}: learning to represent molecules by continuous representations which facilitates the prediction and optimization of the chemical properties of interest; 2) \textit{decoding}: learning to map an optimized continuous representation back to the input molecular graph.

\textbf{Learning representations based on auto-encoding models } 
Many recent studies highlight the applications of generative models for molecule generation tasks, for instance \cite{gomez2018automatic,kusner2017grammar,dai2018syntax,jin2018junction}, just to name a few. One typically chooses to use a VAE \cite{kingma2013auto} for this task, where we wish to learn an encoder for mapping a data point $\textbf{x}$ to a latent code $\textbf{z}$ in the continuous latent space, and a decoder for mapping from $\textbf{z}$ back to $\textbf{x}$. In VAE, the decoder is defined by a likelihood function $p_{\textbf{G}}(\textbf{x}|\textbf{z})$ and parameterized by a neural network $\textbf{G}$. A prior distribution, typically normal distribution $\mathcal{N}(0,\textbf{I})$, is imposed on the latent code $\textbf{z}$ and play the role of a regularizer. The posterior distribution is approximated by a variational distribution $q_{\textbf{E}}(\textbf{z}|\textbf{x})$, which is defined as the encoder and parameterized by another neural network $\textbf{E}$. The VAE simultaneously learns both the parameters of encoder $\textbf{E}$ and decoder $\textbf{G}$ by maximizing the evidence lower bound (ELBO) \cite{kingma2013auto}.

%However, VAE was shown to be hard to train when the decoder is complex \cite{bradshaw2019model,tolstikhin2017wasserstein}. For this, we use the adversarial auto-encoder (AAE) \cite{makhzani2015adversarial} to overcome this limitation of the VAE. AAE employs the same architecture of VAE and regularize the latent space with a discriminator model $\textbf{D}(\textbf{z})$. The discriminator is trained to discriminate samples from the latent distribution and prior distribution, resulting in the following optimization task:
%\begin{equation}
%    \min_{\textbf{E}, \textbf{G}} \max_{\textbf{D}}\mathbb{E}_{\textbf{x}\sim p_{\text{data}}}}\log \textbf{D}\left(\textbf{E}(\textbf{x})\right)+\mathbb{E}_{\textbf{z}\sim p(\textbf{z})}}\log \left(1-\textbf{D}(\textbf{z})\right) -  \mathbb{E}_{\textbf{x}\sim p_{\text{data}}}}\log p\left(\textbf{x}|\textbf{G}\left(\textbf{E}(\textbf{x})\right)\right)
%\end{equation}

\textbf{Generation of chemically valid molecules } SMILES strings \cite{weininger1988smiles} have been widely used to describe the structure of molecules by machine learning (ML) techniques. Prior work on molecule generation tasks formulated the molecule generation as a string generation problem and leverage advances in generative models for text \cite{bowman2015generating} to learn a character variational autoencoder (CVAE, \cite{gomez2018automatic}). However this type of representation is not designed to capture information of structure of molecules, preventing the generative models from generation of valid molecules (i.e. successfully parsed by RDKIT \cite{landrum2006rdkit}). Thus, CVAE often produced chemically invalid molecules. To address the validity problem, several works proposed to use graph representations of molecules such as \cite{jin2018junction, liu2018constrained}. Notably, junction tree variational autoencoder (JT-VAE, \cite{jin2018junction}) proposed to generate molecular graphs with structure-by-structure approach to avoid invalid intermediate states encountered in node-by-node approach, e.g. \cite{li2018learning}, by exploiting the junction tree notion. It was experimentally demonstrated that JT-VAE produces almost 100\% valid molecules when sampled from a prior distribution.

\textbf{Generation of synthesizable molecules } While all of the previous works focus on generation of valid molecules with desired chemical properties, there have been few methods to address the synthesizability of generated molecules. That is how practical it is to make ML-generated molecules from a given set of commercially available reactants (also called starting molecules). For this purpose, Bradshaw et al \cite{bradshaw2019model} proposed to apply a generative model on bags of reactants instead of molecules as in previous approaches. In particular, first, an encoder is to map a set of starting molecules to a continuous vector in the latent space. Second, a decoder is to map the continuous vector back to the set of reactants. As the ultimate goal is to generate molecules, in the generation process, the set of predicted reactants is mapped to a final product molecule through a reaction predictor, named Molecule Transformer \cite{schwaller2019molecular}. 
Thus, this model not only generates a molecule with optimized chemical properties, but also suggest a set of known starting molecules that can be used to synthesize it.
However, a drawback of this model is the limited diversity of generated molecules as it considered only the products as a result of single-step chemical reactions applied on starting molecules, while most natural product molecules are often resulted from multi-step chemical reactions. Concurrent to our work, Bradshaw et al \cite{bradshaw2020barking} proposed to generate molecules via multi-step synthesis and formalizing a multi-step reaction synthesis as a directed acyclic graph (DAG). However, their method was based on an additional reaction predictor, which might generate \emph{imaginary} product molecules. Our method, on the other hand, focuses on chemical reaction trees, for which we generate reaction templates and molecules alternatively.

\section{Proposed method}
In this paper, we tackle two problems of molecular generation task: \emph{generation} of valid molecules with desired chemical properties and \emph{recommendation} of synthetic routes for generated molecules from a set of starting molecules and reaction templates from literature.
As shown in Figure \ref{Fig:chemicalreactiontree}, a synthetic route for a target molecule is described as a tree-like structure, called \emph{reaction tree}, where molecule nodes (circles) and template nodes (rectangles) alternatively appear. The molecule nodes can be either starting molecules (e.g. commercially available molecules) or intermediaries (products of a single-reaction step). The template nodes represent reaction templates, encoded in SMARTS language, which use subgraph matching rules to transform a set of reactants into hypothetical product molecules. These chemical transformations are deterministically done by RDKIT's RunReactants function. Furthermore, the number of reactants can vary for different reaction steps, and can be determined by reaction templates. From a data set of chemical reaction trees, we can extract a vocabulary of starting molecules and another of reaction templates.

Our method focuses on learning to generate reaction trees by alternatively generating their molecule and template nodes. In the generation stage, the generated reaction trees are expected to be valid in the sense that we can perform a number of chemical transformation steps with predicted starting molecules and reaction templates towards the root, where the final product is obtained. However, generation of reaction trees with huge vocabularies of molecules and reaction templates is challenging for standard generative models such as VAEs. Also the majority of reaction templates show up rarely in reaction database \cite{fortunato2020machine}. These challenges prevent the generative models from learning meaningful representations for reaction trees and generating valid synthetic routes in the generation stage.

To deal with the above challenges, our key idea is to couple reaction trees with \emph{junction trees}, used in \cite{jin2018junction}, to represent molecular graphs. A junction tree is a tree structure object, which represents subgraph components and their relative arrangement of a molecular graph. Given a molecule, the junction tree can be constructed using a tree decomposition algorithm (see \cite{jin2018junction} for more details). We argue that using junction trees can benefit the generation of reaction trees because of the two following reasons. First, \emph{junction and reaction trees offer complementary representations of the original molecular graphs}: given information of substructures in junction trees, the generative model can better decide which reaction templates and starting molecules to generate in corresponding reaction trees. Second, \emph{generation of junction trees is easier}: the size of substructure vocabulary, extracted from training set using the tree decomposition algorithm, is much smaller than the numbers of starting molecules and reaction templates in the reaction database. Our generative model extends VAE by using suitable encoders and attention-based decoders for trees in learning representations of both junction and reaction trees. We will detail our model in the following subsections.

\subsection{A Variational Autoencoder For Jointly Learning Junction And Reaction Tree Pairs}
In our model, each molecular graph is associated with a pair of a junction tree and a reaction tree $\left(x,y \right)$, where $x$ denotes the junction tree constructed from the original molecular graph by the tree decomposition, and $y$ denotes the reaction tree constructed by a synthetic planning software, e.g. \cite{chen2020retro}. Our goal is to develop a generative model using both $x$ and $y$, which can improve upon performance of the generative model using $y$ only. 
\begin{wrapfigure}{r}{0.35\textwidth}
  \begin{center}
    \includegraphics[width=0.35\textwidth]{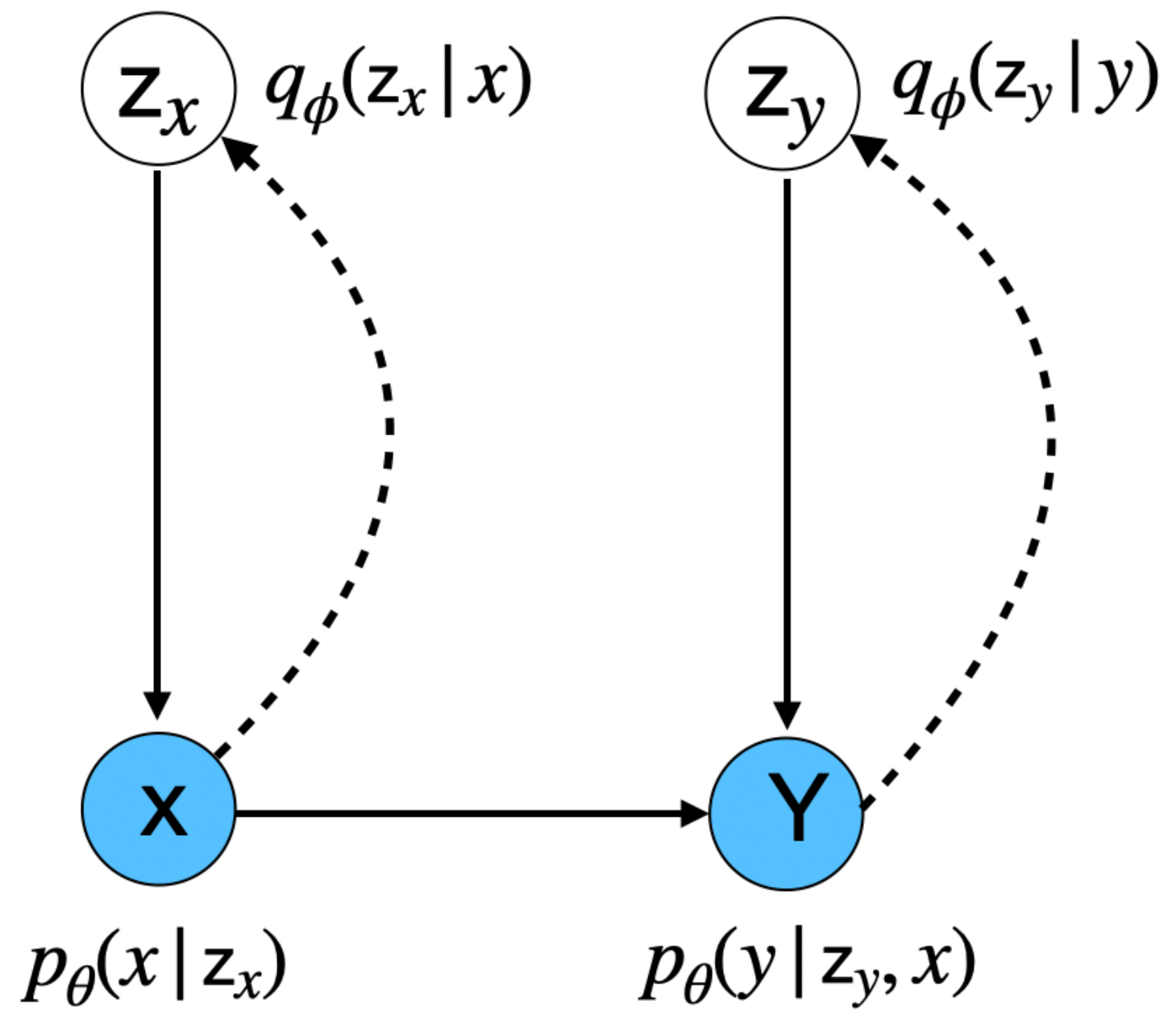}
  \end{center}
  \caption{Probabilistic graphical model of our model. Generative models are denoted by solid lines, and inference models are denoted by dashed lines.}
  \label{Fig:graphicalmodel}
\end{wrapfigure}
\textbf{Generative and Inference Models}. The generative process of a tree pair ($x,y$) is defined as follows: two continuous latent codes $\textbf{z}_{x}, \textbf{z}_{y} \in \mathbb{R}^{d}$ are independently sampled from prior distributions $p(\textbf{z}_{x})$ and $p(\textbf{z}_{y})$. $\textbf{z}_{x}$ is used to generate $x$ through a decoder $p_{\theta}(x|\textbf{z}_{x})$, and $\textbf{z}_{y}$ is used to generate $y$ along with $x$ through another decoder $p_{\theta}(y|\textbf{z}_{y}, x)$. The generative and inference models are illustrated in Figure \ref{Fig:graphicalmodel}. The joint distribution can be factorized as follows:
\begin{equation}
    p_{\theta}(x,y,\textbf{z}_{x},\textbf{z}_{y})=p(\textbf{z}_{x})p(\textbf{z}_{y})p_{\theta}(x|\textbf{z}_{x})p_{\theta}(y|\textbf{z}_{y}, x)
\end{equation}
where $p(\textbf{z}_{x})=p(\textbf{z}_{y})=\mathcal{N}(0,\textbf{I})$ is a Gaussian prior. The variational posterior $q_{\phi}(\textbf{z}_{x},\textbf{z}_{y}|x,y)$ can be trivially factorized as:
\begin{equation}
q_{\phi}(\textbf{z}_{x},\textbf{z}_{y}|x,y)=q_{\phi}(\textbf{z}_{x}|x)q_{\phi}(\textbf{z}_{y}|y)
\end{equation}
where encoders $q_{\phi}(\textbf{z}_{x}|x)$ and $q_{\phi}(\textbf{z}_{y}|y)$ are diagonal Gaussians.

\textbf{Objective}. The variational lower bound of the data likelihood can be derived as follows:
\begin{align}
    \mathcal{L}\left(\theta, \phi, x, y\right)&=  \mathbb{E}_{q_{\phi}(\textbf{z}_{x},\textbf{z}_{y}|x,y)}\log p_{\theta}(x,y|\textbf{z}_{x},\textbf{z}_{y})- \mathrm{KL}\left(q_{\phi}(\textbf{z}_x, \textbf{z}_{y}|x,y)||p(\textbf{z}_{x},\textbf{z}_{y})\right)\nonumber\\
    &= \mathbb{E}_{q_{\phi}(\textbf{z}_{x}|x)}\log p_{\theta}(x|\textbf{z}_{x}) + \mathbb{E}_{q_{\phi}(\textbf{z}_{y}|y)}\log p_{\theta}(y|\textbf{z}_{y},x)\\
    &-\mathrm{KL}\left(q_{\phi}(\textbf{z}_{x}|x)||p(\textbf{z}_{x})\right) -\mathrm{KL}\left(q_{\phi}(\textbf{z}_{y}|y)||p(\textbf{z}_{y})\right)\nonumber
\end{align}
Based on this formulation, our model is decomposed into four main components, each of which is parameterized by a neural network: two \textit{variational neural tree encoders} that model $q_{\phi}(\textbf{z}_{x}|x)$ and $q_{\phi}(\textbf{z}_{y}|y)$, two \textit{variational neural tree decoders} that model $p_{\theta}(x|\textbf{z}_{x})$ and $p_{\theta}(y|\textbf{z}_{y},x)$.

\iffalse
\begin{figure}[h]
\centering
\includegraphics[width=1.0\textwidth]{images/overview_vae.pdf}
 \caption{(a) Example of a synthetic route generated by our proposed model and represented by a chemical reaction tree in which circles and squares represent molecules and reaction template nodes. The template nodes and starting molecule nodes are labeled by their index, while molecule nodes of the root and intermediate products are labeled as -1, indicating more chemical reaction steps are taken from them. The final product molecule ('CC(C)(C)OC(=O)N1CCC[C@H]1C(O)COc1ccccc1') at the root node is obtained by performing chemical transformations from the starting molecule nodes towards the root.
 (b) Depiction of steps taken to generate a chemical reaction tree by our proposed model.}
\label{Fig:reactiontree}
\end{figure}
\fi

\subsection{Encoder and Decoder Networks for Junction Tree: $q_{\phi}(\textbf{z}_{x}|x)$ and $p_{\theta}(x|\textbf{z}_{x})$}
For the reaction tree $x$, we follow encoder and decoder architectures described in \cite{jin2018junction}. 

\textbf{Junction Tree Encoder}. We encode the junction tree $x$ with a tree message passing network. Specifically, each substructure node $i$ is represented by a one-hot encoding $\textbf{x}_{i}$ corresponding to its label type. Each edge $(i,j)$ is associated with two message representations $\textbf{m}_{ij}$ and $\textbf{m}_{ji}$. An arbitrary leaf node is selected as the root and the messages are propagated along the tree in two phases: 1) bottom-up: leaf nodes initialize and propagate messages towards the root; 2) top-down: the root propagate messages down to all the leaf nodes. The message $\textbf{m}_{ij}$ is computed by a Gated Recurrent Unit ($\mathrm{GRU}$, \cite{chung2014empirical}) when all its precursors $\{\textbf{m}_{ki}|k\in \mathcal{N}(i)\setminus j\}$ have been computed as follows: $\textbf{m}_{ij}=\mathrm{GRU}\left(\textbf{x}_{i},\{\textbf{m}_{ki}\mid k\in \mathcal{N}(i)\setminus j\}\right)$.

After two-phase message passing, we obtain a set of node embeddings $\textbf{H}_{x}=\{\textbf{h}_{1},...,\textbf{h}_{|x|}\}$, where $|x|$ denotes the number of nodes in the junction tree $x$; $\textbf{h}_{i}$ is the node embedding of node $i$ and computed from its inward messages via a linear layer, followed by an activation function $g$:
\begin{equation}
    \textbf{h}_{i}=g\left(\textbf{W}^{0}\textbf{x}_{i}+\textbf{U}^{0}\sum_{k\in \mathcal{N}(i)}\textbf{m}_{ki}\right)
    \label{Eqn:nodeembedding}
\end{equation}

We use the node embedding of the root $\textbf{h}_{\text{root}}$ to encode the whole junction tree $x$. The mean $\mu_{x}$ and $\mathrm{log}$ variance $\log \sigma_{x}$ of its variational posterior approximation $q_{\phi}(\textbf{z}_{x}|x)$ are computed from $\textbf{h}_{\text{root}}$ with two separate linear layers. The latent code $\textbf{z}_{\textbf{x}}$ is sampled from a Gaussian $\mathcal{N}(\textbf{z}_{x};\mu_{x}, \sigma_{x})$. 

\textbf{Junction Tree Decoder}. We decode a junction tree from its latent code $\textbf{z}_{x}$ with a tree decoder. The tree is constructed in a top-down fashion from the root by generating nodes in a depth-first order. For every visited node, we first make a topological prediction on whether this node has children to generate. When a new child node is generated, we predict its label and repeat this process. The decoder backtracks when a node has no more child node to generate. We refer the reader to the paper \cite{jin2018junction} for the details of both encoder and decoder for the junction tree.

\subsection{Encoder and Decoder Networks for Reaction Tree: $q_{\phi}(\textbf{z}_{y}|y)$ and $p_{\theta}(y|\textbf{z}_{y},x)$}
\textbf{Reaction Tree Encoder}. Unlike the junction tree encoder, we encode the reaction tree $y$ in the bottom-up fashion from the leaf nodes towards the root. Specifically, each starting molecule node and template node is represented by a one-hot encoding and their representations are computed by performing a lookup of the reaction templates within the template vocabulary and starting molecule within the starting molecule vocabulary. The representations of intermediate molecule nodes are computed when all their reactants' representations have been computed.

Formally, given a single-reaction step with a product molecule node $i$, reaction template $\textit{T}$ and a list of molecule nodes $\{j_{1},...,j_{m}\}$ as reactants, the representation $\textbf{v}_{i}$ of molecule node $i$ is computed by combining the  representations of its reactants $\{\textbf{v}_{j_{1}},...,\textbf{v}_{j_{m}}\}$ and one-hot encoding of reaction template  $\textbf{y}_{\textit{T}}$ via a one hidden layer network, followed by the activation function $\text{g}$:

\begin{equation}
   \textbf{v}_{i}=g\left(\textbf{W}^{1}\textbf{y}_{\textit{T}}+  
   \textbf{U}^{1}\sum_{t=1}^{m}\textbf{v}_{j_{t}}\right)
\end{equation}
We recurse this process until the root is reached. Finally, the representation of the root $\textbf{v}_{\textit{root}}$ (final product molecule) is used as the representation for the entire reaction tree.
The mean $\mu_{y}$ and log variance $\log\sigma_{y}$ of the variational posterior approximation $q_{\phi}(\textbf{z}_{y}|y)$ are computed from $\textbf{v}_{\text{root}}$ with two separate linear layers. The latent code $\textbf{z}_{y}$ is sampled from a Gaussian $\mathcal{N}(\textbf{z}_{y};\mu_{y}, \sigma_{y})$.

\textbf{Reaction Tree Decoder}.
Given the junction tree $x$ and the latent code $\textbf{z}_{y}$, we decode the reaction tree $y$ with a tree decoder. The reaction tree is constructed in the top-down fashion by generating molecule and template nodes alternatively. It is worth noting that, in the generation process, the junction tree $x$ is decoded first and then used to decode the reaction tree $\textbf{y}$. This allows the model to take advantage of information about the substructures of the product molecule that it is going to generate, which benefits the generation of template and molecule nodes in reaction tree $y$. Thus we associate each node $i$ in reaction tree $y$ with a hidden state $\textbf{s}_{i}$ and a context vector $\textbf{c}_{i}$. The context vector $\textbf{c}_{i}$ depends on a set of embeddings $\textbf{H}_{x}=\{\textbf{h}_{1},...,\textbf{h}_{|x|}\}$, which are obtained from $x$ using the junction tree encoder (see Eqn. \ref{Eqn:nodeembedding}). 
\begin{wrapfigure}[22]{r}{0.37\textwidth}
  \begin{center}
    \includegraphics[width=0.35\textwidth]{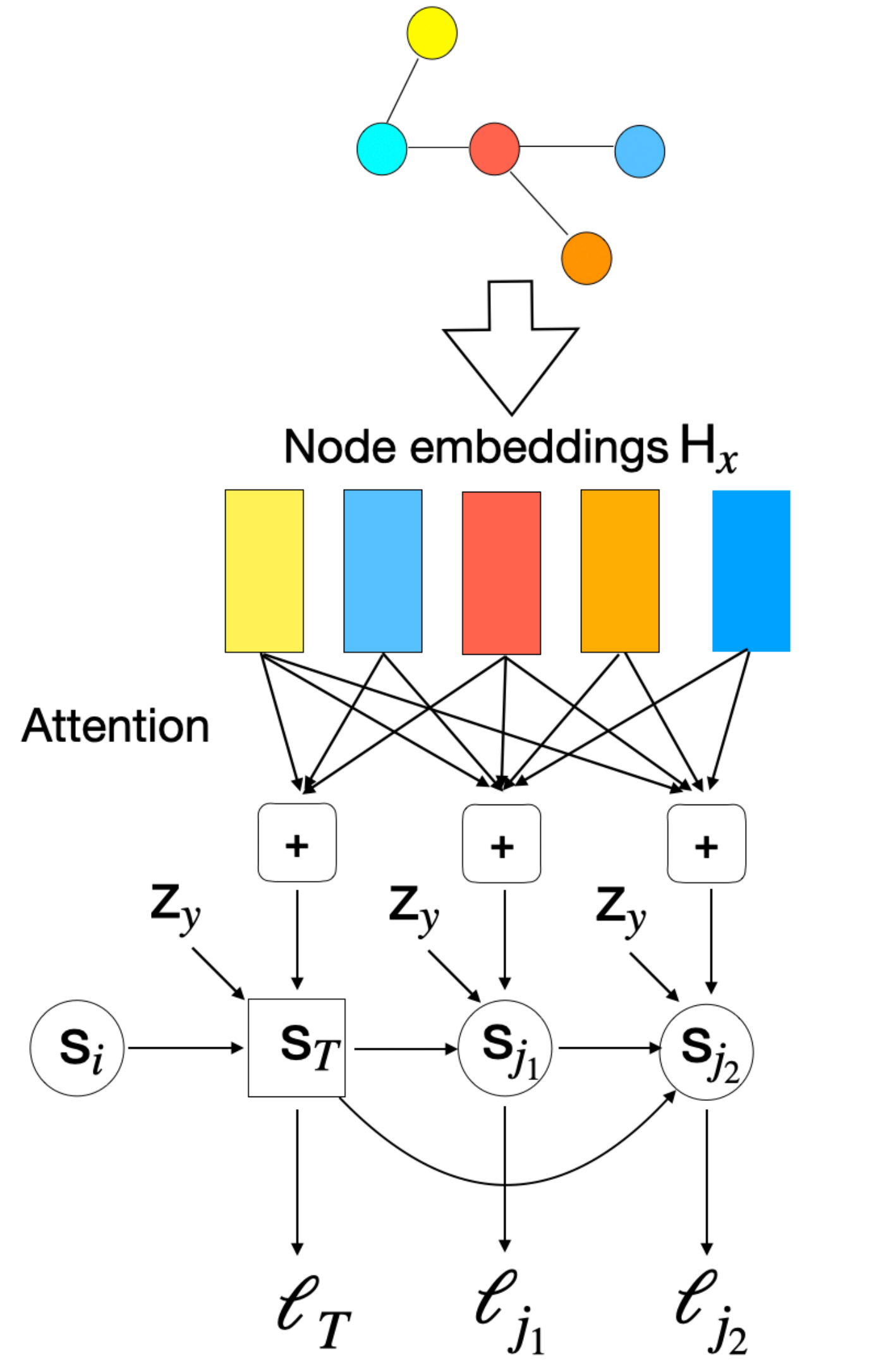}
  \end{center}
  \caption{Depiction of the updating of hidden states by the reaction tree decoder through attention mechanism.}
  \label{Fig:decoder}
\end{wrapfigure}
Formally, the context $\textbf{c}_{i}$ is computed as a weighted sum of these node representations, as follows:
\begin{equation}
    \textbf{c}_{i}=\mathrm{Attention}(\textbf{H}_{x}, \textbf{s}_{i})\coloneqq \sum_{j=1}^{|\textbf{x}|}\alpha_{ij}\textbf{h}_{j} 
\end{equation}
where $\alpha_{ij}=\exp{f(\textbf{s}_{i},\textbf{h}_{j})}/\sum_{k=1}^{|\textbf{x}|}\exp{f(\textbf{s}_{i},\textbf{h}_{k})}$; $f$ is a function which measures how well the hidden state $\textbf{s}_{i}$ matches the node embedding $\textbf{h}_{j}$. In our implementation, we define $f$ as the inner product of two input vectors.

In the following, we will detail the decoding process of the reaction tree. First we generate a hidden state for the root by feeding the latent code $\textbf{z}_{y}$ to a one layer neural network, followed by the activation function $g$: $\textbf{s}_{\text{root}}=g(\textbf{W}^{\textit{root}}\textbf{z}_{y})$. The root is labeled -1, which indicates that the model will perform a reaction step from it.

For every intermediate molecule node $i$, which is labeled -1, with a computed hidden state $\textbf{s}_{i}$, the model generates a template node $T$, computes its hidden state and make a prediction on its label as follows:
\begin{align}
 \textbf{c}_{i}&=\mathrm{Attention}\left( \textbf{H}_{x}, \textbf{s}_{i} \right)\nonumber \\
     \textbf{s}_{T} &= \mathrm{GRU}^{\text{template}}\left( \left[\textbf{z}_{y}, \textbf{c}_{i} \right],\textbf{s}_{i}
     \right)\\
     p_{T} &= \mathrm{softmax}\left( \textbf{W}^{\text{template}} \left[ \textbf{z}_{y}, \textbf{s}_{T} \right]\right)\nonumber
\end{align}
where $\textbf{s}_{T}$ is the hidden state of the generated template node $T$, which is updated from the hidden state of node $i$, the latent code $\textbf{z}_{y}$ and context vector $\textbf{c}_{i}$ through a $\mathrm{GRU}$; $p_{T}$ is the distribution over the vocabulary of templates, computed by combining the latent code $\textbf{z}_{y}$ and hidden state $\textbf{s}_{T}$ through a linear layer, followed by the $\mathrm{softmax}$ function. The label of the template node $\ell_{T}$ is sampled from $p_{T}$.

For each template node $T$, generated from the molecule node $i$, the decoder continues to generate a number of molecule nodes as its reactants $\{j_{1},j_{2},...,j_{m} \}$, where $m$ is the number of reactants of reaction template $T$. The decoder computes their hidden states and predict their labels as follows:
\begin{align}
 \textbf{c}_{j_{k-1}}&=\mathrm{Attention}\left( \textbf{H}_{x}, \textbf{s}_{j_{k-1}} \right)\nonumber\\
 \textbf{s}_{j_{k}} &= \mathrm{GRU}^{\text{molecule}}\left( \left[\textbf{z}_{y}, \textbf{c}_{i}\right],\textbf{s}_{j_{k-1}}\right)\\
p_{j_{k}} &= \mathrm{softmax}\left( \textbf{W}^{\text{molecule}} \left[ \textbf{z}_{y}, \textbf{s}_{j_{k}} \right]\right)\nonumber
\end{align}
where $\textbf{s}_{j_{k}}$ is the hidden state of current reactant molecule node and is computed from the previous reactant's hidden state $\textbf{s}_{j_{k-1}}$ by another $\mathrm{GRU}$ \footnote{For the first reactant ($k=1$), we use the hidden state of the generated template node $T$ to compute the context vector: $\textbf{c}_{j_{0}}=\mathrm{Attention}\left( \textbf{H}_{x}, \textbf{s}_{T} \right)$}; $p_{j_{k}}$ is the distribution over the vocabulary of molecules and predicted by combining the latent code $\textbf{z}_{y}$ and its hidden state $\textbf{s}_{j_{k}}$ via a neural layer, followed by the $\mathrm{softmax}$ function. The label of the molecule node $\ell_{j_{k}}$ is sampled from $p_{j_{k}}$. If $\ell_{j_{k}}$ equals -1, the model will continue to expand the tree. The above process terminates when no more intermediate molecule nodes left. The decoding process of reaction tree is illustrated in Figure \ref{Fig:decoder}.

\subsection{Model Training}
The tree decoders aim to maximize the likelihoods $p_{\theta}(x|\textbf{z}_{x})$ and $p_{\theta}(y|\textbf{z}_{y},x)$ for the junction tree $x$ and reaction tree $y$, respectively. For the junction tree, the decoder minimize the cross entropy loss $\mathcal{L}_{\text{junction}}(x)$ for topology and label prediction of its substructure nodes. As for the reaction tree, the decoder minimize the cross entropy loss $\mathcal{L}_{\text{reaction}}(y)$ for label prediction of its molecule and template nodes. The joint training objective for a training instance $(x,y)$ is defined as: $\mathcal{L}(x,y)=\mathcal{L}_{\text{junction}}(x)+\mathcal{L}_{\text{reaction}}(y)+\mathrm{KL} \text{ loss}$. We optimize the model parameters $\theta$ and $\phi$ jointly using standard gradient descent and the reparameterization trick \cite{kingma2013auto}. The proposed model was implemented with Pytorch and its implementation is available at \url{https://github.com/haidnguyen0909/rxngenerator}.

\section{Experiments}
We demonstrate our proposed model for the task of molecule generation in terms of the three following aspects: \emph{1) Generation of random molecules}: how the model generates molecules when sampling from the prior distribution (see subsection 4.1); \emph{2) Optimization of properties}: how the model generates novel molecules with desired chemical properties by performing search in the latent space with Bayesian optimization (see subsection 4.2); \emph{3) Synthetizability}: how synthesizable the molecules generated by the model are (see subsection 4.3). Below we describe the data and model configuration that we used across the evaluations.

\textbf{Data}. In order to train our model, we need a data set of multi-step chemical reactions. For this, we used \textit{Retro*}, proposed in \cite{chen2020retro}, to generate reaction trees. \textit{Retro*} is a neural-based $A^\star$-like algorithm to find a synthetic route efficiently for a given target molecule. It was shown to outperform previous methods for retrosynthetic planning problem in terms of both success rate and solution quality. We extracted molecules from USPTO reaction data set \cite{lowe2012extraction} and used \textit{Retro*} to synthesize them to obtain a set of (multi-step) chemical reactions. To make sure that starting molecules and reaction templates are popular for the chemists, we filtered out the original set of reactions so that each reaction contains starting molecules and templates that occur at least five times in the filtered set. Furthermore, we found that a number of chemical reactions predicted by \textit{Retro*} was invalid in the sense that the target molecules cannot be synthesized by the predicted synthetic routes. By removing these, we ended up with having a data set of 21218 valid reaction trees, together with vocabularies of 9766 starting molecules and 5567 reaction templates. We also use the tree decomposition to generate a junction tree for each molecule and obtained a vocabulary of 275 valid chemical substructures.

\textbf{Model Configuration}. We set the representation dimension for template and molecule nodes as 200, and the latent space dimension as 50. The model was trained with mini-batch gradient descent with Adam, learning rate of 0.001 and batch size of 32. The activation $g$ we used was $\mathrm{ReLU}$. The model was trained on a NVIDIA TESLA V100 SXM3-32GB with 100 epochs.

\subsection{Generation of random molecules}
We evaluate our model by using the following metrics: validity, uniqueness, novelty, Frechet ChemNet Distance (FCD) and quality, previously used in \cite{bradshaw2019model,jin2018junction,kusner2017grammar,liu2018constrained}. To compute these metrics, we sample 10000 latent codes from the prior $\mathcal{N}(0, \textbf{I})$, and use the trained decoder to decode them into reaction trees. \textit{Validity} is defined as the proportion of valid reaction trees generated by the proposed method. \textit{Uniqueness} is defined as the proportion of valid reaction trees whose product molecules have not been seen before. \textit{Novelty} is defined as the proportion of valid reaction trees whose product molecules are not present in the training set. \textit{Quality} is measured by the proportion of valid reaction trees whose product molecules pass the quality filters proposed by Brown et al \cite{brown2019guacamol}. Finally, we measure \textit{FCD} by computing Wasserstein-2 distance between the product molecules of valid reaction trees and the target molecules of training reaction trees in the training set. 
\begin{table}[ht]
    \centering
    \caption{ Validity, uniqueness, novelty and quality, Quality \cite{brown2019guacamol} (measured in \%, \textit{higher better}) and Frechet ChemNet Distance \cite{preuer2018frechet} (\textit{lower better}) of the product molecules of valid reaction trees by decoding 10000 latent points from the prior $p(\textbf{z})$. The uniqueness, novelty, quality, FCD are conditioned on validity.
    MT stands for the Molecule Transformer \cite{schwaller2019molecular}.
    }
    \begin{tabular}{l c c c  c r}
    \hline
    Model name & Validity & Uniqueness & Novelty & Quality & FCD\\
    \hline
    CVAE \cite{gomez2018automatic} & 12.02 & 56.28 & 85.65 & 52.68 & 37.65\\
    GVAE \cite{kusner2017grammar} & 12.91 & 70.06 & 87.88 & 46.87 & 29.32\\
    CGVAE \cite{liu2018constrained} & 100.0 & 93.51 & 95.88 & 44.45 & 11.73\\
    AAE \cite{kadurin2017cornucopia} & 85.96 & 98.54 & 93.37 & 94.89 & 1.12 \\
    MOLECULE CHEF + MT \cite{bradshaw2019model} & 99.05 & 95.95 & 89.11 & 95.30 & 0.73\\
    DoG-AE \cite{bradshaw2020barking} & 100 & 98.3 & 92.9 & 95.5 & 0.83\\
    Dog-Gen \cite{bradshaw2020barking} & 100 & 97.7 & 98.4 & 101.6 & 0.45\\
    \hline
    Proposed & 64.5 & 73.87 & 66.44 & 95.80 &  0.66 \\
    \hline
    \end{tabular}
    
    \label{tab:metriccomparison}
\end{table}

We compared our model to the following baselines: character VAE (CVAE) \cite{gomez2018automatic}, grammar VAE (GVAE) \cite{kusner2017grammar}, Adversarial autoencoder (AAE) \cite{kadurin2017cornucopia}, constrained graph VAE (CGVAE) \cite{liu2018constrained} and MOLECULE CHEF \cite{bradshaw2019model}. The results are reported in Table \ref{tab:metriccomparison}. We empirically observed that the validity, uniqueness and novelty were around 64\%, 74\% and 66\%, respectively, which are acceptable for the molecule generation task. Also, the proportion of product molecules of valid reaction trees that pass the quality filters was the second highest (quality of 95.80\%) among the compared methods, indicating that molecules resulted from multi-step chemical reactions are rather stable. Interestingly, the proposed method also achieved low \textit{FCD} score.

\begin{figure}[!tbp]
  \centering
  \subfloat{\includegraphics[width=0.5\textwidth]{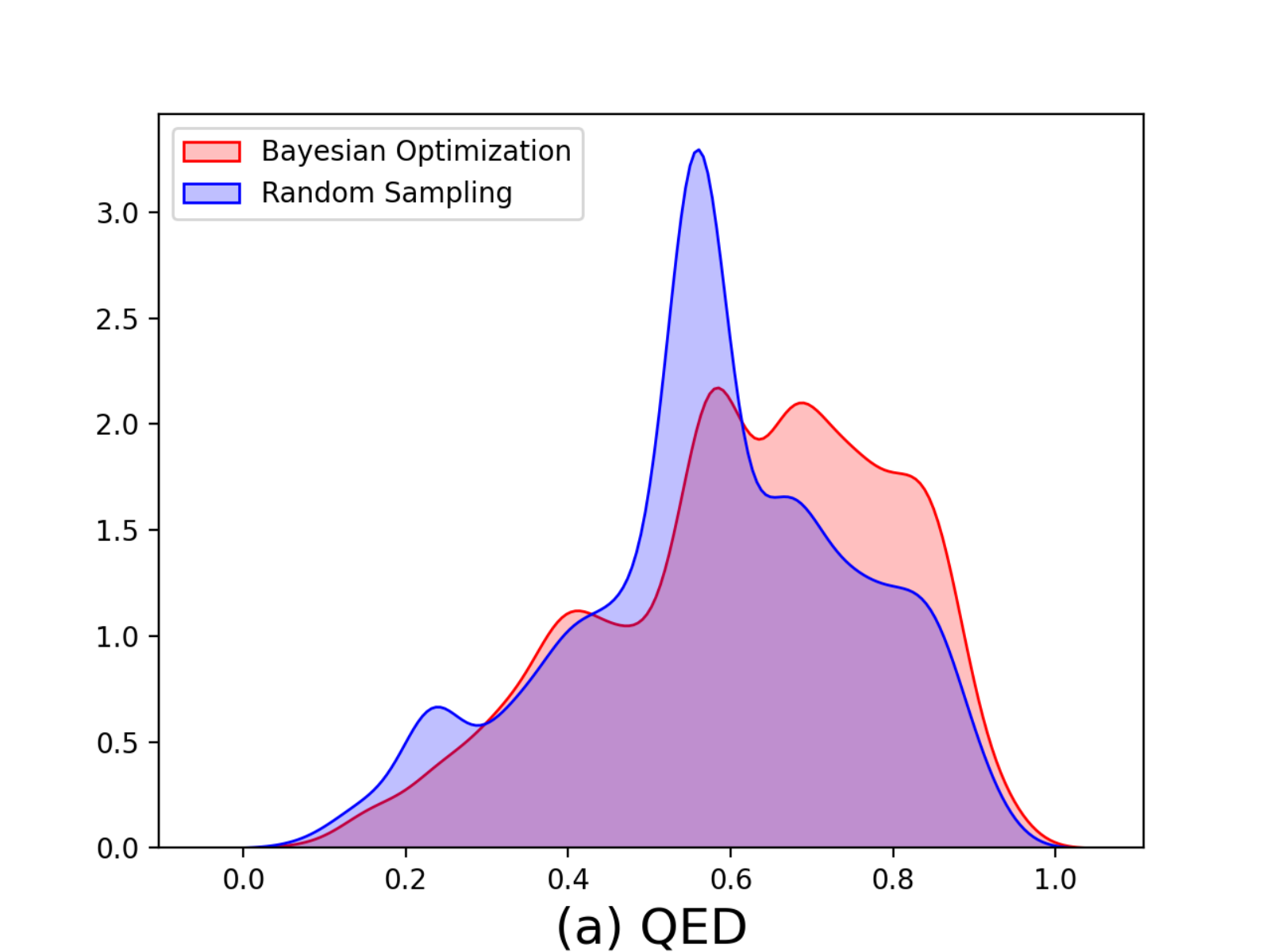}\label{fig:f1}}
  \hfill
  \subfloat{\includegraphics[width=0.5\textwidth]{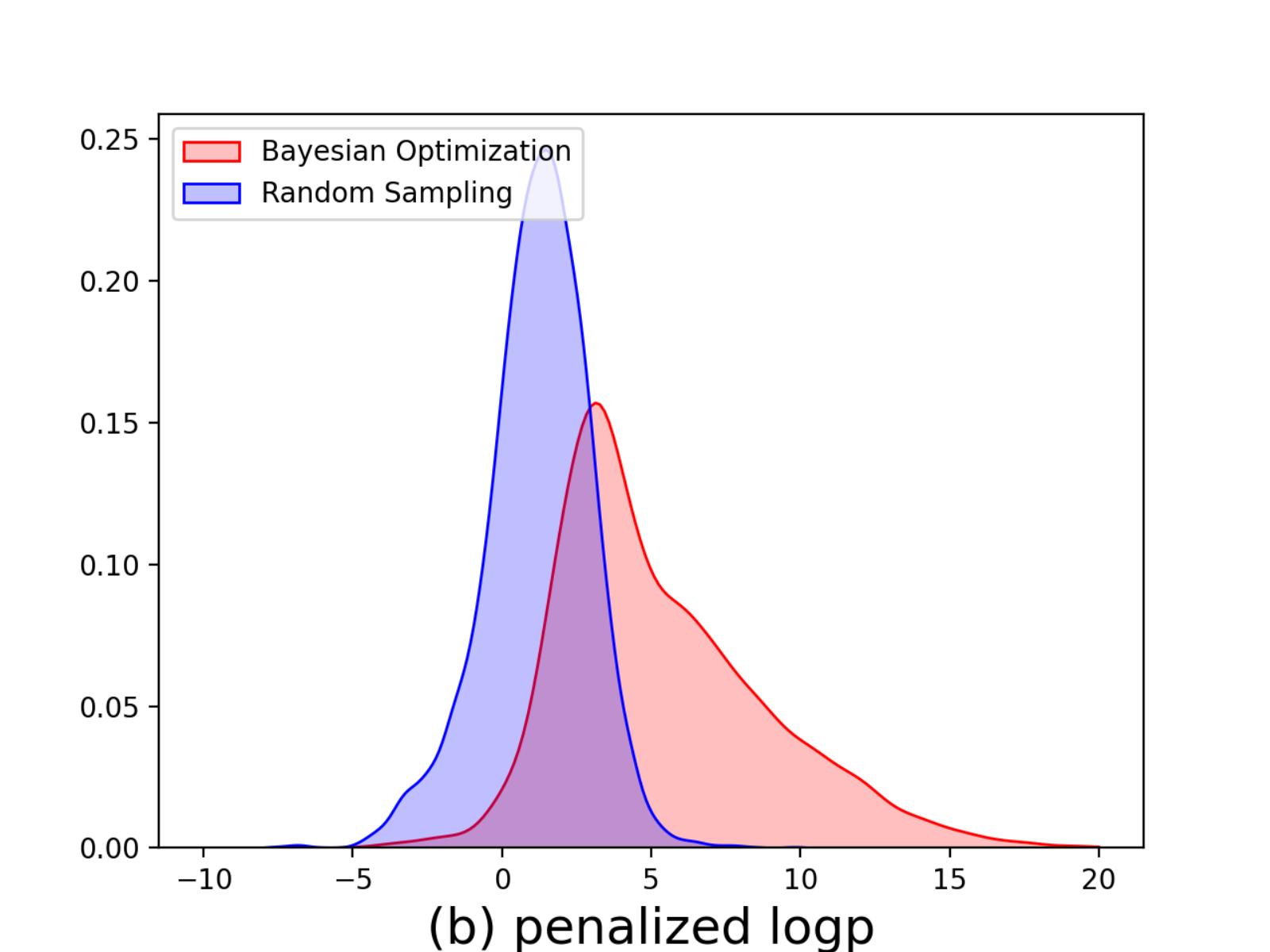}\label{fig:f2}}
  \caption{KDE plots of distributions of (a) QED and (b) penalized logp scores. The distributions of scores obtained by randomly sampling latent vectors from the prior distribution and Bayesian optimization (BO) are shown in blue and red, respectively.
  The plots show that the distributions of scores found by BO have higher mass over higher scores in comparison to scores found by random sampling. The scores are calculated from the product molecules of valid chemical reaction trees generated by the proposed model.}
  \label{fig:kdeplot}
\end{figure}

\begin{figure}[h]
\centering
\includegraphics[width=0.9\textwidth]{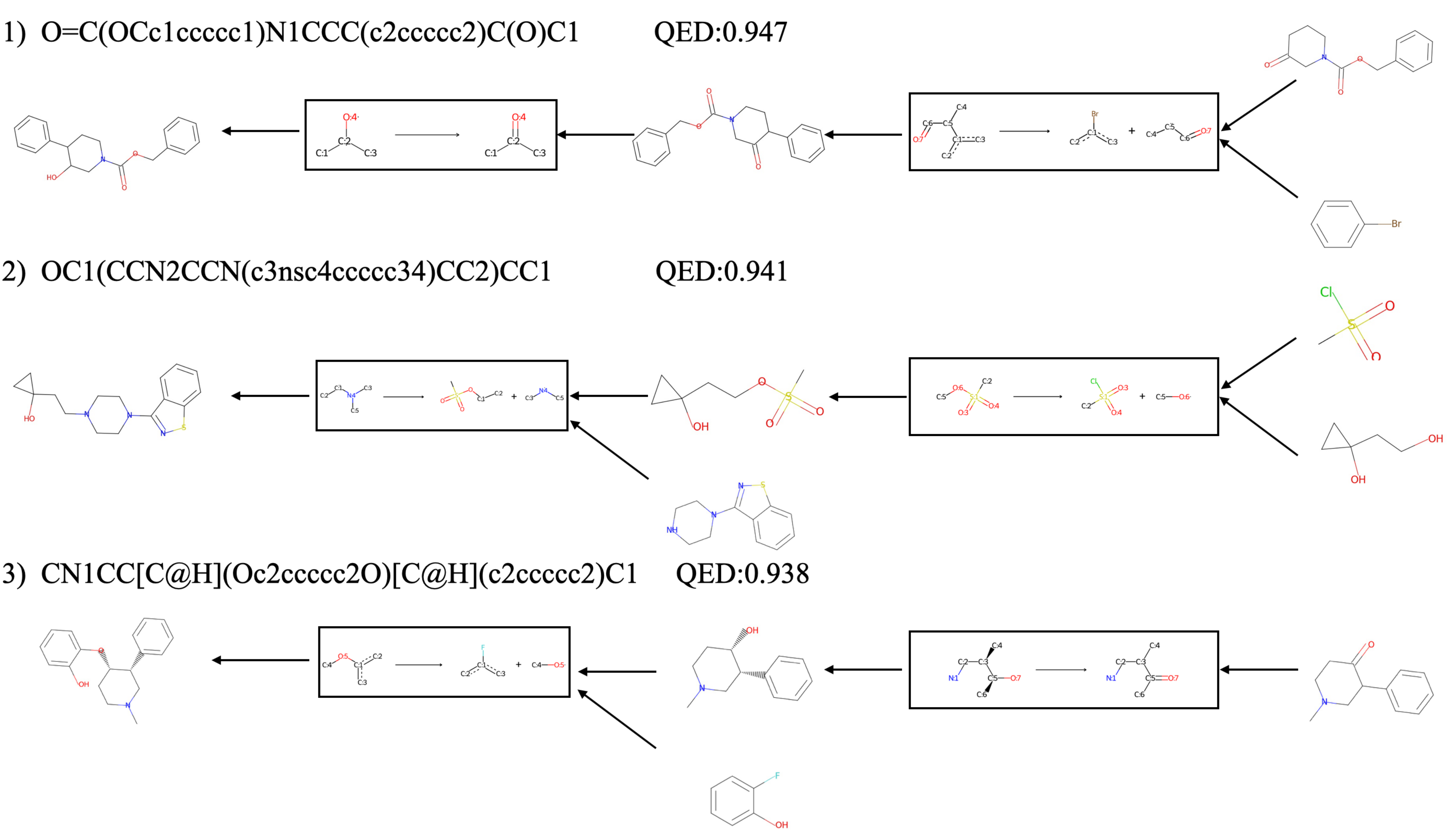}
 \caption{Three examples of reaction trees, found by Bayesian Optimization, have high QED scores.}
 \label{fig:bestreactiontrees}
\end{figure}

\subsection{Search for molecules with desired properties via Bayesian optimization}

We consider finding novel molecules with desired chemical properties. Following the previous work \cite{gomez2018automatic,kusner2017grammar,jin2018junction, dai2018syntax}, we evaluate the two target chemical properties: octanol-water partition coefficients penalized by the synthetic accessibility score and number of long cycles (penalized logP), and Quantitative Estimate of Drug-likeness (QED). We perform Bayesian optimization to optimize for these target properties. We first train our model on the training set of reaction trees, and then obtain a latent code for each reaction tree, given by the mean of the variational encoding distribution. We use sparse Gaussian process (SGP) to predict the score of product molecule synthesized by a reaction tree given its latent code. We perform 5 iterations of batched BO using the expected improvement (EI) and generate 50 new codes in the latent space for each iteration. Each new latent vector is then decoded into new reaction trees whose product molecules are expected to have desired properties.  

As a comparison, we consider product molecules of reaction trees obtained by performing BO in the latent space and those of reaction trees obtained by randomly sampling latent codes from the prior. In Figure \ref{fig:kdeplot}, we plot distributions of (a) QED and (b) penalized logP obtained by BO and random sampling, respectively. As observed, the distributions of scores found through BO have higher mass over higher values in comparison to scores found by random sampling, showing the effectiveness of BO in searching novel molecules with desired properties in the latent space. 

We also compared the top-3 molecules with best scores: QED and penalized logP found by BO under different models. We compare our results to those of the following baselines: CVAE \cite{gomez2018automatic}, GVAE \cite{kusner2017grammar}, SD-VAE \cite{dai2018syntax} and JT-VAE \cite{jin2018junction}. As shown in Table \ref{tab:top3scores}, our method cound find molecules with better scores than compared methods in both QED and penalized logP. We show the product molecules of reaction trees with the best QED scores in Figure \ref{fig:bestreactiontrees}.

\begin{table}[ht]
\caption{Best property scores: QED and penalized logP, found by each method. We compared to the results of the following baselines: CVAE \cite{gomez2018automatic}, GVAE \cite{kusner2017grammar}, SD-VAE \cite{dai2018syntax} and JT-VAE \cite{jin2018junction}}
\begin{center}
\begin{tabular}{l c c c c c r}
    \hline
    \multirow{2}{*}{Method} & \multicolumn{3}{c}{Penalized logP} & \multicolumn{3}{c}{QED}\\

    & $1^{\text{st}}$ & $2^{\text{nd}}$ & $3^{\text{rd}}$ & $1^{\text{st}}$ & $2^{\text{nd}}$ & $3^{\text{rd}}$\\
    \hline
    CVAE \cite{gomez2018automatic} & 1.98 & 1.42 & 1.19 & - & - & -\\
    GVAE \cite{kusner2017grammar} & 2.94 & 2.89 & 2.80 & - & - & - \\
    SD-VAE \cite{dai2018syntax} & 4.04 & 3.50 & 2.96 & - & - & -\\
    JT-VAE \cite{jin2018junction} & 5.30 & 4.93 & 4.49 & 0.947 & 0.947 & 0.945\\
    \hline
    Proposed & 26.19 & 20.71 & 20.36 & 0.947 & 0.947 & 0.0.941\\
    \hline
\end{tabular}
\end{center}
\label{tab:top3scores}
\end{table}

\begin{table}[ht]
\caption{Comparison of \textit{synthetizability rates} of molecules generated by the following models: JT-VAE \cite{jin2018junction}, MOLECULE CHEF \cite{bradshaw2019model}, DoG-Gen \cite{bradshaw2020barking} and the proposed model.}
\begin{center}
\begin{tabular}{l c c c r}
    \hline
    \textbf{Methods} & JT-VAE \cite{jin2018junction} & MOLECULE CHEF \cite{bradshaw2019model}& DoG-Gen \cite{bradshaw2020barking} & Proposed\\
    \hline
     \textit{Synthetizability rate}(\%) & 50.5 & 65.6 & 67.1 & \textbf{97.81}\\
     \hline
\end{tabular}
\end{center}
\label{tab:synthesis_rate}
\end{table}

\subsection{The synthetizability of generated molecules by models}
The novel feature of our model is its ability to learn to decode a reaction tree instead of a molecule from a code in the latent space, from which we can produce a product molecule by performing chemical transformation steps. Thus, we argue that our proposed model generates molecules with high synthesizability.  In this section, we compare the synthetizability of molecules generated by different models.

We compared our method to JT-VAE \cite{jin2018junction}, MOLECULE CHEF \cite{bradshaw2019model} and DoG-Gen \cite{bradshaw2020barking}. For each of the comparing methods, we sampled 1000 latent codes from the prior, and then decoded each 10 times, selected the most frequent generated molecule as a representative. Ideally, we have 1000 molecules generated by each method. We used \textit{Retro*} \cite{chen2020retro} to synthesize the generated molecules. We measured \textit{synthetisizability rate (\%)}, which is defined as the proportion of generated molecules that can be successfully synthesized by \textit{Retro*}.

We show the \textit{synthetisizability rate} in Table \ref{tab:synthesis_rate}. We empirically observed that a half of molecules generated by JT-VAE cannot be synthesized by \textit{Retro*}. MOLECULE CHEF also had a similar \textit{synthetisizability rate} (65.6\%) as it was restricted to molecules which are products of singe-step chemical reactions. DoG-Gen had a slightly better rate (67\%).
Our method achieved the highest rate by directly learning to generate full synthetic routes with specific reaction templates and molecules. 

\section{Conclusion}
In this work, we have proposed a generative model for generating multi-step chemical reactions by exploiting the junction tree notion. The proposed model tackles two problems of molecule generation: 1) generation of valid molecules with desired chemical properties and 2) synthesizability of the generated molecules. Experiments showed that the model could generate chemical reactions whose product molecules are with high chemical properties. More importantly, each generated molecule is associated with a complete synthetic route, making it practical to synthesize in 
reality. one of the main limitations of our method is the need of synthetic routes of training molecules associated with huge numbers of starting molecules and reaction templates, which are not always available or limited in practice. Our future work would be a focus on the development of machine learning models which can be used with both reaction trees and also molecules with unknown synthetic routes.

\textbf{Broader Impact}

Bringing a new drug to market is money- and time-consuming process with the average cost of over one billion USD and time of 13 years from discovery to market. Machine learning (or more broadly Artificial Intelligence (AI)) is having an increasing impact on accelerating many parts of the drug discovery pipeline. The current approaches to drug molecular generation leads to most molecules being unstable or impossible to synthesize in reality. We hope that our proposed model with the ability to produce synthesizable molecules, contributes to the research in this direction.

Besides the positive impact that AI could bring for faster drug discovery, it is also important for us to be aware of possible negative aspects. For instance, these AI technologies could be possible for faster development of chemical substances whose toxic properties are used to kill, injure or incapacitate human beings. Another negative impact of AI based molecular generation technologies is their increased automation process, which might limit our insights of the molecular design process. We think that in order to reduce the possibilities of these risks, it is necessary to have interactions and conversations between chemists and those who are working on machine learning to be clear about the goal of the research. Also the development of explainable machine learning models could be more emphasized to help better understand the ML decision making process.

\medskip

\bibliographystyle{plainnat}
\bibliography{reference}

\begin{thebibliography}{22}
\providecommand{\natexlab}[1]{#1}
\providecommand{\url}[1]{\texttt{#1}}
\expandafter\ifx\csname urlstyle\endcsname\relax
  \providecommand{\doi}[1]{doi: #1}\else
  \providecommand{\doi}{doi: \begingroup \urlstyle{rm}\Url}\fi

\bibitem[Bowman et~al.(2015)Bowman, Vilnis, Vinyals, Dai, Jozefowicz, and
  Bengio]{bowman2015generating}
Samuel~R Bowman, Luke Vilnis, Oriol Vinyals, Andrew~M Dai, Rafal Jozefowicz,
  and Samy Bengio.
\newblock Generating sentences from a continuous space.
\newblock \emph{arXiv preprint arXiv:1511.06349}, 2015.

\bibitem[Bradshaw et~al.(2019)Bradshaw, Paige, Kusner, Segler, and
  Hern{\'a}ndez-Lobato]{bradshaw2019model}
John Bradshaw, Brooks Paige, Matt~J Kusner, Marwin~HS Segler, and
  Jos{\'e}~Miguel Hern{\'a}ndez-Lobato.
\newblock A model to search for synthesizable molecules.
\newblock \emph{arXiv preprint arXiv:1906.05221}, 2019.

\bibitem[Bradshaw et~al.(2020)Bradshaw, Paige, Kusner, Segler, and
  Hern{\'a}ndez-Lobato]{bradshaw2020barking}
John Bradshaw, Brooks Paige, Matt~J Kusner, Marwin~HS Segler, and
  Jos{\'e}~Miguel Hern{\'a}ndez-Lobato.
\newblock Barking up the right tree: an approach to search over molecule
  synthesis dags.
\newblock \emph{arXiv preprint arXiv:2012.11522}, 2020.

\bibitem[Brown et~al.(2019)Brown, Fiscato, Segler, and
  Vaucher]{brown2019guacamol}
Nathan Brown, Marco Fiscato, Marwin~HS Segler, and Alain~C Vaucher.
\newblock Guacamol: benchmarking models for de novo molecular design.
\newblock \emph{Journal of chemical information and modeling}, 59\penalty0
  (3):\penalty0 1096--1108, 2019.

\bibitem[Chen et~al.(2020)Chen, Li, Dai, and Song]{chen2020retro}
Binghong Chen, Chengtao Li, Hanjun Dai, and Le~Song.
\newblock Retro*: learning retrosynthetic planning with neural guided a*
  search.
\newblock In \emph{International Conference on Machine Learning}, pages
  1608--1616. PMLR, 2020.

\bibitem[Chung et~al.(2014)Chung, Gulcehre, Cho, and
  Bengio]{chung2014empirical}
Junyoung Chung, Caglar Gulcehre, KyungHyun Cho, and Yoshua Bengio.
\newblock Empirical evaluation of gated recurrent neural networks on sequence
  modeling.
\newblock \emph{arXiv preprint arXiv:1412.3555}, 2014.

\bibitem[Dai et~al.(2018)Dai, Tian, Dai, Skiena, and Song]{dai2018syntax}
Hanjun Dai, Yingtao Tian, Bo~Dai, Steven Skiena, and Le~Song.
\newblock Syntax-directed variational autoencoder for structured data.
\newblock \emph{arXiv preprint arXiv:1802.08786}, 2018.

\bibitem[Fortunato et~al.(2020)Fortunato, Coley, Barnes, and
  Jensen]{fortunato2020machine}
Michael~E Fortunato, Connor~W Coley, Brian~C Barnes, and Klavs~F Jensen.
\newblock Machine learned prediction of reaction template applicability for
  data-driven retrosynthetic predictions of energetic materials.
\newblock In \emph{AIP Conference Proceedings}, volume 2272, page 070014. AIP
  Publishing LLC, 2020.

\bibitem[G{\'o}mez-Bombarelli et~al.(2018)G{\'o}mez-Bombarelli, Wei, Duvenaud,
  Hern{\'a}ndez-Lobato, S{\'a}nchez-Lengeling, Sheberla, Aguilera-Iparraguirre,
  Hirzel, Adams, and Aspuru-Guzik]{gomez2018automatic}
Rafael G{\'o}mez-Bombarelli, Jennifer~N Wei, David Duvenaud, Jos{\'e}~Miguel
  Hern{\'a}ndez-Lobato, Benjam{\'\i}n S{\'a}nchez-Lengeling, Dennis Sheberla,
  Jorge Aguilera-Iparraguirre, Timothy~D Hirzel, Ryan~P Adams, and Al{\'a}n
  Aspuru-Guzik.
\newblock Automatic chemical design using a data-driven continuous
  representation of molecules.
\newblock \emph{ACS central science}, 4\penalty0 (2):\penalty0 268--276, 2018.

\bibitem[Goodfellow et~al.(2014)Goodfellow, Pouget-Abadie, Mirza, Xu,
  Warde-Farley, Ozair, Courville, and Bengio]{goodfellow2014generative}
Ian~J Goodfellow, Jean Pouget-Abadie, Mehdi Mirza, Bing Xu, David Warde-Farley,
  Sherjil Ozair, Aaron Courville, and Yoshua Bengio.
\newblock Generative adversarial networks.
\newblock \emph{arXiv preprint arXiv:1406.2661}, 2014.

\bibitem[Jin et~al.(2018)Jin, Barzilay, and Jaakkola]{jin2018junction}
Wengong Jin, Regina Barzilay, and Tommi Jaakkola.
\newblock Junction tree variational autoencoder for molecular graph generation.
\newblock In \emph{International Conference on Machine Learning}, pages
  2323--2332. PMLR, 2018.

\bibitem[Kadurin et~al.(2017)Kadurin, Aliper, Kazennov, Mamoshina, Vanhaelen,
  Khrabrov, and Zhavoronkov]{kadurin2017cornucopia}
Artur Kadurin, Alexander Aliper, Andrey Kazennov, Polina Mamoshina, Quentin
  Vanhaelen, Kuzma Khrabrov, and Alex Zhavoronkov.
\newblock The cornucopia of meaningful leads: Applying deep adversarial
  autoencoders for new molecule development in oncology.
\newblock \emph{Oncotarget}, 8\penalty0 (7):\penalty0 10883, 2017.

\bibitem[Kingma and Welling(2013)]{kingma2013auto}
Diederik~P Kingma and Max Welling.
\newblock Auto-encoding variational bayes.
\newblock \emph{arXiv preprint arXiv:1312.6114}, 2013.

\bibitem[Kusner et~al.(2017)Kusner, Paige, and
  Hern{\'a}ndez-Lobato]{kusner2017grammar}
Matt~J Kusner, Brooks Paige, and Jos{\'e}~Miguel Hern{\'a}ndez-Lobato.
\newblock Grammar variational autoencoder.
\newblock In \emph{International Conference on Machine Learning}, pages
  1945--1954. PMLR, 2017.

\bibitem[Landrum et~al.(2006)]{landrum2006rdkit}
Greg Landrum et~al.
\newblock Rdkit: Open-source cheminformatics.
\newblock 2006.

\bibitem[Li et~al.(2018)Li, Vinyals, Dyer, Pascanu, and
  Battaglia]{li2018learning}
Yujia Li, Oriol Vinyals, Chris Dyer, Razvan Pascanu, and Peter Battaglia.
\newblock Learning deep generative models of graphs.
\newblock \emph{arXiv preprint arXiv:1803.03324}, 2018.

\bibitem[Liu et~al.(2018)Liu, Allamanis, Brockschmidt, and
  Gaunt]{liu2018constrained}
Qi~Liu, Miltiadis Allamanis, Marc Brockschmidt, and Alexander~L Gaunt.
\newblock Constrained graph variational autoencoders for molecule design.
\newblock \emph{arXiv preprint arXiv:1805.09076}, 2018.

\bibitem[Lowe(2012)]{lowe2012extraction}
Daniel~Mark Lowe.
\newblock \emph{Extraction of chemical structures and reactions from the
  literature}.
\newblock PhD thesis, University of Cambridge, 2012.

\bibitem[Preuer et~al.(2018)Preuer, Renz, Unterthiner, Hochreiter, and
  Klambauer]{preuer2018frechet}
Kristina Preuer, Philipp Renz, Thomas Unterthiner, Sepp Hochreiter, and Günter
  Klambauer.
\newblock Fr{\'e}chet chemnet distance: a metric for generative models for
  molecules in drug discovery.
\newblock \emph{Journal of chemical information and modeling}, 58\penalty0
  (9):\penalty0 1736--1741, 2018.

\bibitem[Schwaller et~al.(2019)Schwaller, Laino, Gaudin, Bolgar, Hunter, Bekas,
  and Lee]{schwaller2019molecular}
Philippe Schwaller, Teodoro Laino, Th{\'e}ophile Gaudin, Peter Bolgar,
  Christopher~A Hunter, Costas Bekas, and Alpha~A Lee.
\newblock Molecular transformer: a model for uncertainty-calibrated chemical
  reaction prediction.
\newblock \emph{ACS central science}, 5\penalty0 (9):\penalty0 1572--1583,
  2019.

\bibitem[Shibukawa et~al.(2020)Shibukawa, Ishida, Yoshizoe, Wasa, Takasu,
  Okuno, Terayama, and Tsuda]{shibukawa2020compret}
Ryosuke Shibukawa, Shoichi Ishida, Kazuki Yoshizoe, Kunihiro Wasa, Kiyosei
  Takasu, Yasushi Okuno, Kei Terayama, and Koji Tsuda.
\newblock Compret: a comprehensive recommendation framework for chemical
  synthesis planning with algorithmic enumeration.
\newblock \emph{Journal of cheminformatics}, 12\penalty0 (1):\penalty0 1--14,
  2020.

\bibitem[Weininger(1988)]{weininger1988smiles}
David Weininger.
\newblock Smiles, a chemical language and information system. 1. introduction
  to methodology and encoding rules.
\newblock \emph{Journal of chemical information and computer sciences},
  28\penalty0 (1):\penalty0 31--36, 1988.

\end{thebibliography}
%%%%%%%%%%%%%%%%%%%%%%%%%%%%%%%%%%%%%%%%%%%%%%%%%%%%%%%%%%%%
\section*{Checklist}
\iffalse
%%% BEGIN INSTRUCTIONS %%%
The checklist follows the references.  Please
read the checklist guidelines carefully for information on how to answer these
questions.  For each question, change the default \answerTODO{} to \answerYes{},
\answerNo{}, or \answerNA{}.  You are strongly encouraged to include a {\bf
justification to your answer}, either by referencing the appropriate section of
your paper or providing a brief inline description.  For example:
\begin{itemize}
  \item Did you include the license to the code and datasets? \answerYes{See Section~\ref{gen_inst}.}
  \item Did you include the license to the code and datasets? \answerNo{The code and the data are proprietary.}
  \item Did you include the license to the code and datasets? \answerNA{}
\end{itemize}
Please do not modify the questions and only use the provided macros for your
answers.  Note that the Checklist section does not count towards the page
limit.  In your paper, please delete this instructions block and only keep the
Checklist section heading above along with the questions/answers below.
%%% END INSTRUCTIONS %%%
\fi
\begin{enumerate}

\item For all authors...
\begin{enumerate}
  \item Do the main claims made in the abstract and introduction accurately reflect the paper's contributions and scope?
    \answerYes{}
  \item Did you describe the limitations of your work?
    \answerYes{}
  \item Did you discuss any potential negative societal impacts of your work?
    \answerYes{}
  \item Have you read the ethics review guidelines and ensured that your paper conforms to them?
    \answerYes{}
\end{enumerate}

\item If you are including theoretical results...
\begin{enumerate}
  \item Did you state the full set of assumptions of all theoretical results?
    \answerNA{}
	\item Did you include complete proofs of all theoretical results?
    \answerNA{}
\end{enumerate}

\item If you ran experiments...
\begin{enumerate}
  \item Did you include the code, data, and instructions needed to reproduce the main experimental results (either in the supplemental material or as a URL)?
    \answerYes{See Section 4.}
  \item Did you specify all the training details (e.g., data splits, hyperparameters, how they were chosen)?
     \answerYes{See Section 4.}
	\item Did you report error bars (e.g., with respect to the random seed after running experiments multiple times)?
   \answerNo{We ran experiment once.}
	\item Did you include the total amount of compute and the type of resources used (e.g., type of GPUs, internal cluster, or cloud provider)?
    \answerYes{See Section 4.}
\end{enumerate}

\item If you are using existing assets (e.g., code, data, models) or curating/releasing new assets...
\begin{enumerate}
  \item If your work uses existing assets, did you cite the creators?
    \answerNA{}
  \item Did you mention the license of the assets?
    \answerNA{}
  \item Did you include any new assets either in the supplemental material or as a URL?
    \answerNA{}
  \item Did you discuss whether and how consent was obtained from people whose data you're using/curating?
    \answerNA{}
  \item Did you discuss whether the data you are using/curating contains personally identifiable information or offensive content?
   \answerNA{}
\end{enumerate}

\item If you used crowdsourcing or conducted research with human subjects...
\begin{enumerate}
  \item Did you include the full text of instructions given to participants and screenshots, if applicable?
    \answerNA{}
  \item Did you describe any potential participant risks, with links to Institutional Review Board (IRB) approvals, if applicable?
    \answerNA{}
  \item Did you include the estimated hourly wage paid to participants and the total amount spent on participant compensation?
   \answerNA{}
\end{enumerate}

\end{enumerate}

%%%%%%%%%%%%%%%%%%%%%%%%%%%%%%%%%%%%%%%%%%%%%%%%%%%%%%%%%%%%

%\appendix

%\section{Appendix}

%Optionally include extra information (complete proofs, additional experiments and plots) in the appendix.
%This section will often be part of the supplemental material.

\end{document}